\documentclass{article}

\usepackage[utf8]{inputenc}

\usepackage{graphicx}
\usepackage{amsmath}
\usepackage{amsthm}
\usepackage{verbatim}
\usepackage{xspace}
\usepackage{subfigure}
\usepackage{booktabs}
\usepackage{microtype}
\usepackage{hyperref}
\usepackage{bbold}

\usepackage[accepted]{icml2018}

\newtheorem*{example*}{Example}

\newcommand\Mapper{\textsc{Mapper}\xspace}
\newcommand\FiFa{\textsc{FiFa}\xspace}

\icmltitlerunning{Fibres of Failure}

\begin{document}
\twocolumn[
\icmltitle{Fibres of Failure: Classifying errors in predictive processes}

\begin{icmlauthorlist}
\icmlauthor{Leo Carlsson}{KTH}
\icmlauthor{Gunnar Carlsson}{ayasdi,stanford}
\icmlauthor{Mikael Vejdemo-Johansson}{csi}
\end{icmlauthorlist}

\icmlaffiliation{KTH}{KTH Royal Institute of Technology}
\icmlaffiliation{ayasdi}{Ayasdi Inc.}
\icmlaffiliation{stanford}{Stanford University}
\icmlaffiliation{csi}{CUNY College of Staten Island}

\icmlcorrespondingauthor{Leo Carlsson}{leoc@kth.se}
\icmlcorrespondingauthor{Mikael Vejdemo-Johansson}{mvj@math.csi.cuny.edu}

\vskip 0.3in
] 

\printAffiliationsAndNotice{}

\begin{abstract}
  We describe Fibres of Failure (\FiFa), a method to classify failure modes of predictive processes using the \Mapper algorithm from Topological Data Analysis. 
  Our method uses \Mapper to build a graph model of input data stratified by prediction error.
  Groupings found in high-error regions of the \Mapper model then provide distinct failure modes of the predictive process.

  We demonstrate \FiFa on misclassifications of MNIST images with added noise, and demonstrate two ways to use the failure mode classification: either to produce a correction layer that adjusts predictions by similarity to the failure modes; or to inspect members of the failure modes to illustrate and investigate what characterizes each failure mode.

\end{abstract}

 \section{Introduction}
\label{sec:introduction}



In recent years the interest in transparent, interpretable and explainable models in machine learning has grown dramatically, with dedicated workshops at NIPS 2016\cite{IMLCS-16}, NIPS 2017 \cite{TIML-17, IML-17} and ICML 2017 \cite{WHI-17} as well as attention from grant agencies \cite{XAI}.

The approaches to interpretable models go in several distinct directions -- producing sparse models \cite{hara_finding_2016,wisdom_interpretable_2016,hayete_mdl-motivated_2016,tansey_interpretable_2017}, visualization techniques \cite{smilkov_embedding_2016,selvaraju_grad-cam:_2016,thiagarajan_treeview:_2016,gallego-ortiz_interpreting_2016,krause_using_2016,zrihem_visualizing_2016,handler_visualizing_2016}, hybrid models \cite{krakovna_increasing_2016,reing_toward_2016}, input data segmentation \cite{samek_interpreting_2016,hechtlinger_interpretation_2016,thiagarajan_treeview:_2016}, and model diagnostics with or without blackbox interpretation layers \cite{lundberg_unexpected_2016,vidovic_feature_2016,whitmore_mapping_2016,ribeiro_nothing_2016,singh_programs_2016,phillips_interpretable_2017,ribeiro_model-agnostic_2016,ribeiro_why_2016} to name a few prominent directions.

In this paper, we present a method, \emph{Fibres of Failure} that draws on topological data analysis to produce model diagnostics through a classification of prediction failure modes in feature space. Our method relates to both the input data segmentation and the model diagnostics directions of research by finding and classifying input regions that behave unexpectedly or erroneously as compared to what the model is designed to predict.

Noisy input as well as adversarial learning has been used to motivate and to generate examples and insights for interpretability \cite{kindermans_investigating_2016}. We will use the same basic idea to illustrate our method -- by studying prediction failures on MNIST images with added noise.


\section{Related work}
\label{sec:related-work}



One interpretability method with a large impact on the field, LIME \cite{ribeiro_why_2016}, inspects single instances by perturbing the input and tracing how predictions change with the perturbation. Other interpretability methods focus closer on aggregates of inputs, such as TreeView \cite{thiagarajan_treeview:_2016}, which visualizes deep neural networks by first clustering neurons by activation patterns, then clusters these groups by prediction labels, and finally trains a predictor to predict the meta-clusters from the input data directly.

The \FiFa method builds on \Mapper, an algorithm from Topological Data Analysis that constructs a graph (or simplicial complex) model of arbitrary data. \Mapper has had success in a wide range of application areas, from medical research studying cancer, diabetes, asthma and many more topics \cite{BreastCancer,diabetes,asthma,schneider}, genetics and phenotype studies \cite{fragilex,opinion,camara,TopSoil,RNAhairpin}, to hyperspectral imaging, material science, sports and politics \cite{Lille1,Lille2,NanoMat,lum_extracting_2013}. Of note for our approach are in particular the contributions on cancer, diabetes and fragile X syndrom \cite{BreastCancer,fragilex,diabetes} where \Mapper was used to extract new subgroups from a segmentation of the input space.

Our results build on two fundamental concepts: viewing predictive models as functions and therefore usable as input to \Mapper, and the \Mapper technique for producing intrinsic graph models of arbitrary data sets. 

As a running illustration in this paper we will be looking at how a CNN trained on the MNIST dataset fails when encountering noisy images derived from MNIST. The influence of noise on learning algorithm performance has been studied. \cite{DistortedImagesCNN, distortion2} found a dramatic increase in error rates with increased image distortion, confirming our choice of illustrative test case. Adversarial learning is another method that has been proven successful at deteriorating performance for trained networks \cite{Houdini,attackDefenses,deepfool,PoisoningAttacks}.

A lot of work has been done on making deep networks more robust against perturbations: both against noise deterioration and against adversarial manipulation \cite{robustness,formalGuarentees,RegularizingDNNs,parseval,strongAdversary,ensembleAdversarial}.

\section{Proposed method}
\label{sec:proposed-model}

The proposed method, \textsc{Fibres of Failure (FiFa)}, takes a different approach from the related work. We do not intend to modify deep neural network models, rather we create classifiers on top of the model that recognizes specific types of faulty predictions (failure modes) from a deep learning model trained to recognize MNIST images.

\subsection{\Mapper}
\label{sec:mapper}


\Mapper \cite{singh_topological_2007} is an algorithm that constructs a graph (more generally a simplicial complex) model for a point cloud data set.  The graph is constructed systematically from some well defined input data.  It was defined in \cite{singh_topological_2007}, and has been shown to have great utility in the study of various kinds of data sets (as described in Section \ref{sec:related-work}). It can be viewed as a method of unsupervised analysis of data, in the same way as principal component analysis, multidimensional scaling, and projection pursuit can, but it is more flexible than any of these methods.  Comparisons of the method  with standard methods in the context of hyperspectral imaging have been documented in \cite{Lille1,Lille2}.  

In topological language, \Mapper starts with the choice of a collection of continuous \emph{filter functions} and an open cover over their range. The \emph{fibres}, or preimages of this open cover produces an open cover on the data space, which can be refined using connected components. Doing this with a fine enough cover and non-degenerate filter functions produces a good cover in the sense of the nerve lemma~\cite{hatcher2002algebraic}, so the nerve complex is homotopy equivalent with the data source.

An open cover here is almost, but not quite the same thing as a partition. In order to track connectivity information, the partition cannot be allowed to become disconnected -- that would miss parts of the space, and introduce artificial disconnects. The open cover most cleanly translates into a ``fattened'' partition, or a partition with overlaps between adjacent parts. 

In more detail, and using a more data-focused and less topological description, \Mapper proceeds by the following steps. We let $X$ (the dataset) be a finite metric space.

\begin{enumerate}
\item Select arbitrary functions $f_1,f_2, \ldots , f_k:X \rightarrow \mathbb{R}$. We call these \emph{filter functions} and they encode a separation of datapoints. In practice, the number $k$ is usually $1,2,$ or $3$.  Common filter functions are statistically meaningful quantities such as the values of a density estimator or centrality measure, or outputs from a machine learning algorithm such as PCA or MDS, or a variable used in defining the data set.
\item For each of the functions, pick parameters to produce an overlapping partition of $\mathbb{R}$: a number $N_i$ of partitions and a proportion of overlap $0<p<1$.
\item For each function $f_i$, let $a_i$ and $b_i$ denote the minimum and maximum values taken by $f_i$, and construct an open cover of the interval $J^i=[a_i, b_i]$ by introducing $N_i$ subintervals 
\[
J_s^i = [a_i + (s-1-p/2)\Delta_i, a_i + (s+p/2)\Delta_i]\subseteq J^i
\]
 where $\Delta_i=(b_i-a_i)/N_i$ and $1 \leq s \leq N_i$.
\item Construct a (likely overlapping) partition of $X$ by letting each 
\[
U_{s_1,\dots,s_k} = f_1^{-1}(J^1_{s_1}) \cap \dots \cap f_k^{-1}(J^k_{s_k})
\]
with $1\leq s_i\leq N_i$ be a part in in the partition. 
\item Apply some clustering algorithm to each $U_{s_1,\dots,s_k}$ to decompose it into disjoint sets $U_{s_1,\dots,s_k;j}$. For our experiments, we use single linkage clustering with a heuristic for cutoff based on the histogram of distances in $U$, described in detail in \cite{singh_topological_2007}.
\item Construct a graph by setting the vertices to $\sigma = (s_1,\dots,s_k;j)$ and connecting $\sigma,\dots,$ to $\sigma'$ with anedge precisely when 
\[
U_{\sigma}\cap U_{\sigma'} \neq\emptyset
\]
For the simplicial complex version, vertices $[\sigma_0,\dots,\sigma_d]$ are connected when their joint intersection is non-empty.
\end{enumerate}

  See \cite{singh_topological_2007,carlsson_topology_2009} for more details. 

\Mapper has several implementations available: Python Mapper \cite{pymapper}, Kepler Mapper \cite{kepler-mapper} and TDAmapper\footnote{\url{http://cran.r-project.org/web/packages/TDAmapper}} are all open source, while Ayasdi Inc.\footnote{\url{http://ayasdi.com}} provides a commercial implementation of the algorithm.
For our work we are using the Ayasdi implementation of \Mapper.

\subsubsection{Mapper on prediction failure}
\label{sec:mapp-pred-fail}

The \emph{filters} in the \Mapper function have the effect of ensuring separation of features in the data that are separated by the filter functions themselves.
Step one of \FiFa specifically uses a \Mapper analysis with \emph{prediction error} as one of the filter functions.
By including prediction error this way, the \FiFa algorithm \emph{guarantees} that any groups that are extracted are homogenous with respect to prediction failure, and thus useable as a failure mode designation.

We name a \Mapper model with prediction failure as a filter a \emph{\FiFa model}.

\subsection{Extract subgroups}
\label{sec:extract-subgroups}

Subgroups of the \FiFa model with tight connectivity in the graph structure and with homogenous and large average prediction failure per component cluster provide a classification of failure modes. These can be selected either manually, or using a community detection algorithm. 

When selecting failure modes manually, a visualization such as in Figure~\ref{fig:NetworkCNN} is most helpful. Here, flares (tightly connected subgraphs emanating from a core, such as Group 40) or tightly connected components, loosely connected to surrounding parts of the graph, are the most compelling characterizations of a good failure mode subgroup.


\subsection{Quantitative: model correction layer}
\label{sec:quant-model-corr}

Once failure modes have been identified, one way to use the identification is to add a correction layer to the predictive process. Use a classifier to recognize input data similar to a known failure mode, and adjust the predictive process output according to the behavior of the failure mode in available training data.

\subsubsection{Train classifiers}
\label{sec:train-classifiers}

For our illustrative examples, we demonstrate several ``one vs rest'' binary classifier ensembles where each classifier is trained to recognize one of the failure modes (extracted subgroups) from the Mapper graph. We demonstrate performance of \FiFa for model correction using Linear SVM, Logistic Regression, and Naïve Bayes classifiers.

\subsubsection{Evaluate bias}
\label{sec:evaluate-bias}

A classifier trained on a failure mode may well capture larger parts of test data than expected. As long as the space identified as a failure mode has consistent bias, it remains useful for model correction: by evaluating the bias in data captured by a failure mode classifier we can calibrate the correction layer.

\subsubsection{Adjust model}
\label{sec:adjust-model}

The actual correction on new data is a type of ensemble model, and has flexibility on how to reconcile the bias prediction with the original model prediction -- or even how to reconcile several bias predictions with each other. For our example in this paper we choose to override the CNN prediction with the observed ground truth in the failure mode from the training data used to create the classifier. For regression tasks we have also used the average of the failure mode training group as an offset to subtract from the model prediction.

\subsection{Qualitative: model inspection}
\label{sec:qual-model-insp}

Identifying distinct failure modes and giving examples of these is valuable for model inspection and debugging. Statistical methods, such as Kolmogorov-Smirnov testing, can provide measures of how influential any one feature is in distinguishing one group from another and can give notions of what characterizes any one failure mode from other parts of input space. With examples and distinguishing features in hand, we can go back to the original model design and evaluate how to adapt the model to handle the failure modes better.

Much of the work in interpretability for machine learning provides tools to inspect examples, and for providing a model explanation for a specific example. These work well in conjunction with \FiFa to find explanations for the identified failure modes.

\section{Experiments}
\label{sec:examples}

In order to evaluate the \FiFa method we have trained a CNN classifier on the MNIST data set, created prediction failures by adding noise to the data, and gone through the \FiFa pipeline for the resulting erroneous predictions. With distinct failure modes extracted, we then illustrate both a quantitative and a qualitative approach to handling the output from \FiFa: on the one hand we adjust predictions using classifiers trained on recognizing each failure mode and measure the improvement in classification on the resulting ensemble approach, on the other hand we compare several failure modes that misclassify versions of the same digit (the digit 5) in different ways.

\subsection{Protocol}
\label{sec:protocol}

\begin{figure}
\includegraphics[width=1.0\linewidth]{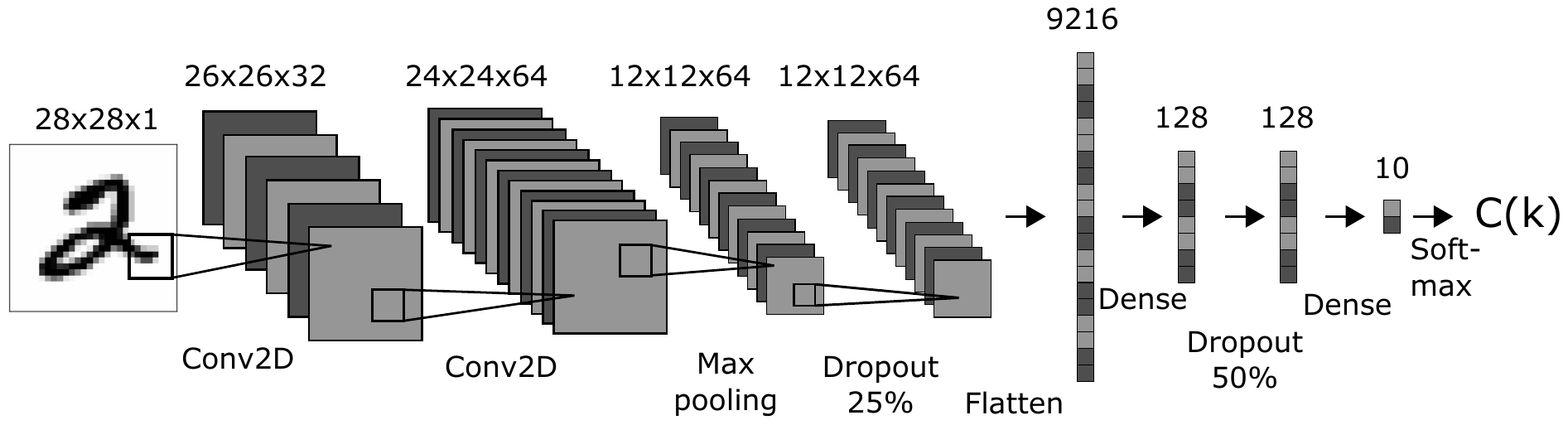}
\caption{The topology for the CNN model. The numbers display the dimension of each layer in the model. The abbreviations, such as Conv2D, describes the specific transformations performed between layers in the model. The activation functions for the classification layer was 'Softmax' and for the other layers 'ReLU'. The optimizer used was 'Adadelta'.}
\label{fig:CNNmodel}
\end{figure}

We created a CNN model with a topology shown in Figure \ref{fig:CNNmodel}. The network topology and parameters was chosen arbitrarily with the only condition that it performs well on the original MNIST data set. The activation functions was 'Softmax' for the classification layer and 'ReLU' for all other layers. The optimizer was Adadelta with $learning rate=1.0$, $\rho=0.95$, and $\epsilon=1e-7$. We trained the model on 60,000 clean MNIST training images and tested it on 10,000 clean MNIST images through 12 epochs. The accuracy on the test-set of 10,000 clean MNIST images was 99.05\%. We created 10,000 corrupt MNIST images using 25\% random binary flips on the clean test images[source for code]. The accuracy on the corrupt MNIST images was 40.45\%.

To create the \Mapper graph we used the following:

\textbf{Filters:} Principal Component 1, probability of Predicted digit, probability of Ground truth digit, and Ground truth digit. Our measure of predictive error is the probability of Ground truth digit. By including the Ground truth digit itself we separate the model on ground truth, guaranteeing that any one one failure mode has a consistent ground truth that can be used for corrections.

\textbf{Metric:} Variance Normalized Euclidean

\textbf{Variables:} 9472 network activations: all activations after the Dropout layer that finishes the convolutional part in the network and before the softmax layer that provides the final predictions. These are the layers with 9216, 128 and 128 nodes displayed in Figure~\ref{fig:CNNmodel}.

\textbf{Instances:} We randomly shuffled the data from the 10,000 clean and 10,000 corrupt images that were used to test the CNN model, and split the 20,000 instances into 5 training sets of size 16,000 each and 5 test sets of size 4,000 each. The training sets was used to create 5 \Mapper graphs. This is in order to perform 5-fold cross validation on the classifiers.

The use of probabilities for predicted and ground truth digit as filters guarantees that \Mapper separates regions of correct predictions from those of wrong predictions. After all, these probabilities are measures of error for the CNN model. We purposely omitted the activations from the Dense-10 layer as input variables because of the direct reference to the probabilities for both the ground truth digit and the predicted digit. 

The following variables were included in the analysis but were not used to create the \FiFa model:

\textbf{10 activations} from the Dense-10 layer, which consists of the probabilities for each digit, 0-9.

\textbf{784 pixel values} representing the flattened MNIST image of size 28x28x1.

\textbf{6 variables}: prediction by the CNN model, ground truth digit, corrupt or original data (binary), correct or incorrect prediction(binary), probability of the Predicted digit (highest value of the Dense-10 layer), and probability of ground truth digit.

Hence, the total number of variables in our analysis were 10272.

To extract failure modes from the \FiFa model we used a supervised community detection method to find groups of approximately constant prediction error. In the \Mapper implementation we are using a grouping method based on Agglomerative Hierarchical Clustering (AHCL)~\cite{edwards1965method,murtagh2012algorithms} and Louvain Modularity\cite{Louvain} is included. As supervision, a function on the data is chosen -- for \FiFa, choose the measure of prediction error. The difference in means of the supervision function produces a graph edge weighting: edges are weighted as ``strong'' if they have similar supervision function values, and ``weak'' if the supervision function values are different. With the graph weighting in place, hierarchical clustering produces a clustering tree using the weighted edges to generate a graph metric to cluster over. Finally, Louvain modularity identifies an optimal graph partition from the clustering tree.

From partitioned groups, we retain as failure modes those groups that have at least 15 data points and have less than 99.05\% correct predictions, which is the accuracy of the CNN model on the original MNIST test data.

We trained classifiers in a one vs. rest scheme on each group in the 5 folds of data that were used to create the 5 \Mapper graphs. We used the following types of classifiers with varying parameters shown in square brackets:

\textbf{Linear-SVM} Loss function: squared hinge, Penalty function: $\ell_2$. Regularization parameter, $C=[0.001, 0.01, 0.1, 1, 10, 100, 1000]$.

\textbf{Logistic Regression} Penalty function: $\ell_2$. Regularization parameter, $C=[0.001, 0.01, 0.1, 1, 100, 10, 1000]$.  

\textbf{Naïve Bayes} Gaussian Naïve Bayes using class priors for each group in the training data set.

We used the parameters from each best performing classifier to train new models. This time, we evaluated each model on second test data set, called 'Corrupt', which consisted of 10,000 new corrupt images using 25\% binary flips on the original MNIST test dataset. Hence, we used the same noise setup as the corrupt images used for testing the CNN model. 

For the test data sets, we evaluated to what extent each classifier predicted member points with the same ground truth digit as that of the group the model was trained on. As we trained the classifiers on groups containing a lot of wrong predictions, it is expected that the classifiers will classify member points with wrong predictions on the test data sets. Hence, we offset the predicted digits with the ground truth digit of the group it was trained on. We attempt to exploit the consistent bias of the classifiers to improve the accuracy of the now combined CNN and classifier ensemble.

\subsection{Quantitative Results}
\label{sec:quantitative-results}

The following parameters were chosen for the three classifiers we evaluated as model correction layers:

\textbf{Linear-SVM} $C=1$. (chosen as highest accuracy in a 5-fold crossvalidation)

\textbf{Logistic Regression} $C=1000$ (chosen as highest accuracy in a 5-fold crossvalidation)

\textbf{Naïve Bayes} Gaussian Naïve Bayes using priors induced from data.

The average number of data points in all failure mode groups in the 5 folds were 4937 of the total 16,000. The average number of clean data points in all groups in the 5 folds were 10.4, accounting for a fraction of 0.21\% of the 4937 data points. This also means that the failure mode groups encompasses roughly 62\% of all corrupt data points in the training set. The number of failure modes (extracted subgroups) in each fold were 41, 41, 41, 41, and 37, respectively.


\begin{figure}[tb]
\includegraphics[width=1.0\linewidth]{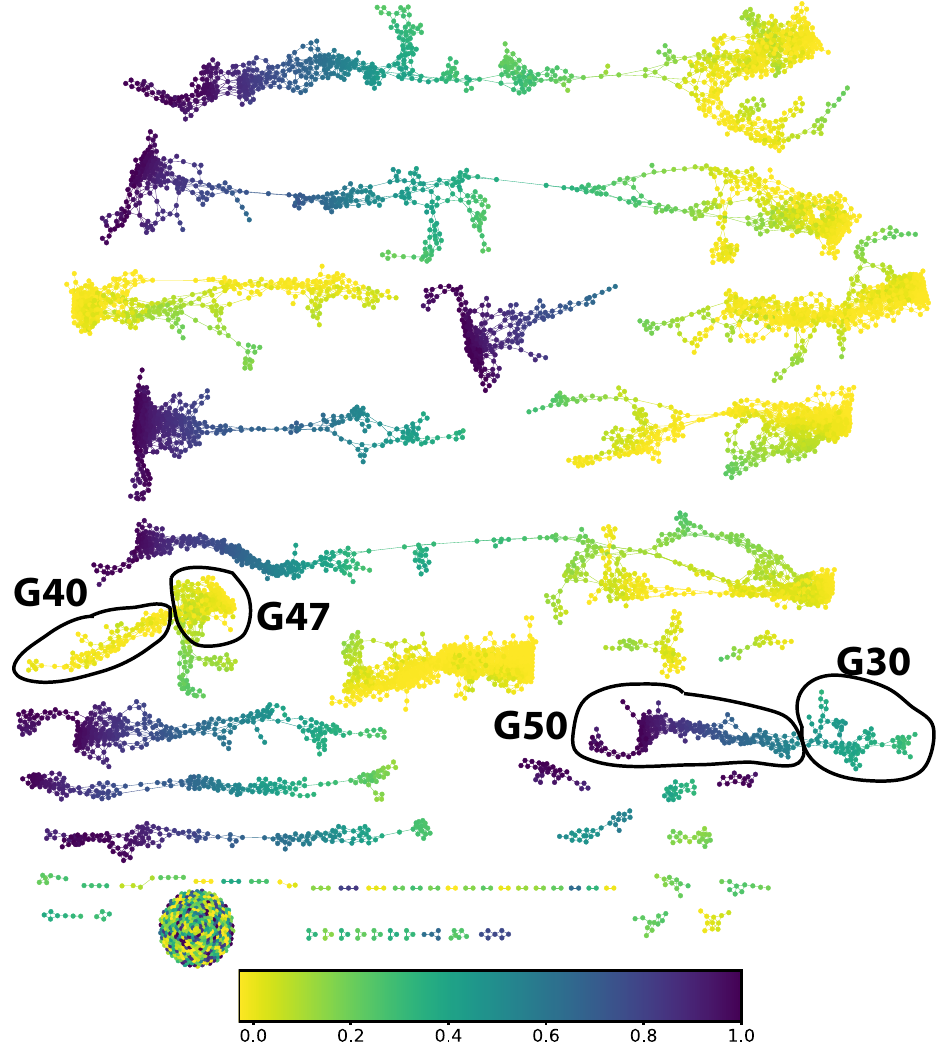}
\caption{The Mapper graph for the CNN on MNIST dataset colored with probability of predicting the ground truth digit. The colorbar is for interpreting the values of the coloring. The circled nodes and edges are the groups Group30, Group40, Group47, and Group50. The 5-fold Mapper graphs are shown in the Supplement. }
\label{fig:NetworkCNN}

\end{figure}


Table \ref{tab:MNISTpreds} shows the accuracy on the two test data sets using CNN with and without \textsc{FiFa}. The linear-SVM classifiers performed best on both data sets with an improvement by 6.43\%pt on the 5-fold cross validation test sets and 19.33\%pt on the 'Corrupt data'.


\begin{table}[tb]
  \begin{tabular}{llll}
    \toprule
    & \textbf{5-fold} & \% clean &  \textbf{Corrupt}  \\ 
      & (4\,000) & 49.7\% &   (10\,000)   \\ \midrule
     CNN & 69.40\% & 48.53\%  & 41.14\%  \\ \midrule \midrule
     CNN+LR & 75.45\%  & \textbf{0.21\%} & 59.57\%  \\
    CNN+SVM & \textbf{75.83\%} & 0.32\% & \textbf{60.47\%}    \\
     CNN+NB & 73.33\% & 6.85\% & 48.34\% \\
     
    \bottomrule
  \end{tabular}
  \caption{Performance of the CNN as compared to CNN with \FiFa driven improvements both on the average of the 5 folds of test data and on entirely corrupted test data. The improvements by each classifier ensemble are for the best performing parameters. In addition to prediction accuracy, we also report the average proportion of uncorrupted (clean) data points in the 5-fold test data set as well as in the predicted data points by each classifier ensemble. Bold face marks the best performances (highest accuracy; lowest percentage of clean digits caught). Noticeably linear classifiers perform well, producing an almost 20\%pt increase in accuracy on corrupted data while imposing corrections on almost no clean images.}
  \label{tab:MNISTpreds}
\end{table}

\subsection{Qualitative Results}
\label{sec:qualitative-results}

\begin{figure}
\includegraphics[width=1.0\linewidth]{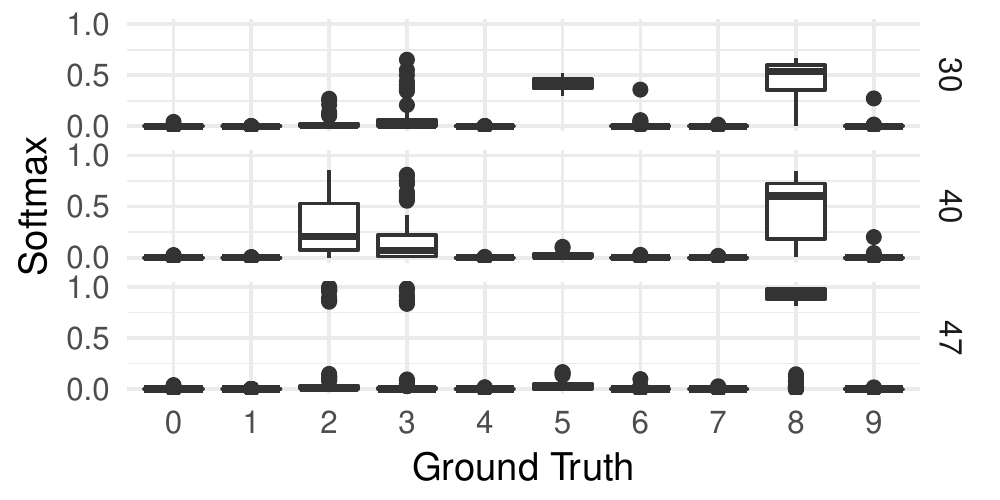}
\caption{The failure modes for a ground truth of 5. We see the distributions of predictions for the three failure modes: only group 30 attaches any significant likelihood to the digit 5 at all, while all three favor 8. For group 40, the digits 2 and 3 are also commonly suggested, while this happens somewhat more rarely in groups 30 and 47.}
\label{fig:Group40distribution}

\end{figure}

For the qualitative analysis, we chose to focus on four groups with digit 5 as the ground truth digit. Group 50, which is not one of the failure mode groups and Groups 30, 40, and 47, all part of the total 39 failure mode groups. The locations of each group are shown in Figure \ref{fig:NetworkCNN}. The distribution of predicted probabilities for each label is shown in Figure \ref{fig:Group40distribution}: group 30 is the group with highest probability to the digit 5, while 40 and 47 are more focused on 8, 2, and 3. All three groups favor digit 8 as their mean probabilities are between 0.5-0.9. 

We compared these three failure modes with the non-failure Group 50 and extracted the 5 activations with the highest KS-values from the Dense-128 layer. See Figure \ref{fig:CNNmodel}. To illustrate the differences between the three failure modes regarding the activations, we have provided a selection of saliency maps~\cite{simonyan2013deep} for all images considered as true members of each of the three failure mode groups. These were all produced using the \texttt{keras-vis} Python package.

\begin{figure}
Activation 24 -- one row each for groups 30, 40 and 47 \\
\includegraphics{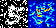}
\includegraphics{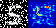}
\includegraphics{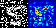}
\includegraphics{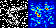} \\
\includegraphics{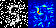}
\includegraphics{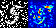}
\includegraphics{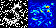}
\includegraphics{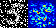} \\
\includegraphics{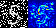}
\includegraphics{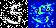}
\includegraphics{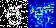}
\includegraphics{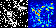} \\
Activation 81 -- one row each for groups 30, 40 and 47 \\
\includegraphics{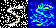}
\includegraphics{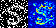}
\includegraphics{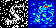}
\includegraphics{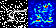} \\
\includegraphics{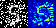}
\includegraphics{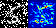}
\includegraphics{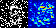}
\includegraphics{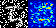} \\
\includegraphics{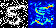}
\includegraphics{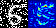}
\includegraphics{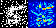}
\includegraphics{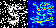} \\
\\ Group 30, activation 89 \\
\includegraphics{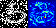}
\includegraphics{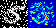}
\includegraphics{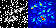}
\includegraphics{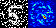} 
\\ Group 30, activation 124 \\
\includegraphics{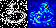}
\includegraphics{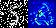}
\includegraphics{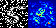}
\includegraphics{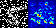} 
\\ Group 40, activation 89 \\
\includegraphics{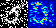}
\includegraphics{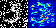}
\includegraphics{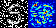}
\includegraphics{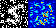} 
\\ Group 40, activation 99 \\
\includegraphics{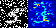}
\includegraphics{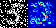}
\includegraphics{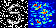}
\includegraphics{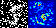} 
\\ Group 40, activation 119 \\
\includegraphics{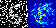}
\includegraphics{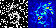}
\includegraphics{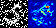}
\includegraphics{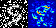} 
\\ Group 47, activation 89 \\
\includegraphics{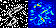}
\includegraphics{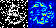}
\includegraphics{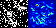}
\includegraphics{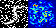} 
\\ Group 47, activation 122 \\
\includegraphics{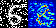}
\includegraphics{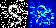}
\includegraphics{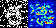}
\includegraphics{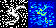} 
\caption{Example noisy images and saliency maps for activations in the penultimate dense layer for the three main failure modes identified for noisy 5s. The two leftmost images were chosen as the most clear saliency maps with respect to digits. The two rightmost were selected based on unclear/noisy saliency maps. All saliency maps are from images classified as members of the respective failure mode group.}
\label{fig:false5images}
\end{figure}

Figure~\ref{fig:false5images} shows a selection of noisy images and their saliency maps for some of the activations highest KS-values within the Dense-128 layer. The two leftmost image pairs were selected based on visual clear saliency maps with respect to digits. The two rightmost were selected based on most unclear/noisy saliency maps. The full collection of saliency maps for these groups can be found in our supplemental material. 

The activations 24 and 81, present in all three groups, display activity that is consistent with an activation detecting features of the digit 5, while the activations 89 and 99 correspond closer to an activation for the digit 3 and 119, 122 and 124 correspond to activations for the digit 8. In particular in the last three groups, noise that closes loops in a written 5 tend to have high saliency.

In Table~\ref{tab:blankSaliency} we show the percentage of blank saliency maps, indicating that an activation is missing completely for a particular input.

\begin{table}[tb]
  \begin{tabular}{llllll}
    \toprule
    \textbf{Group 30} &  & & &  &    \\ \midrule
     Neuron & \textbf{24} & \textbf{33}  & \textbf{81} & 89 & 124 \\ 
     \%Blank & 36.1\%  & 26.2\% & 60.7\% & 0\%  &  8.2\%\\
    \bottomrule
    \textbf{Group 40} &  &  &  &  &    \\ \midrule
     Neuron & \textbf{24} & \textbf{81}  & 89 & 99 & 119 \\ 
     \%Blank & 82.2\%  & 91.8\% & 0\% & 4.1\%  &  17.9\%\\
    \bottomrule
    \textbf{Group 47} &  & & &  &    \\ \midrule
     Neuron & \textbf{24} & \textbf{49}  & \textbf{81} & 89 & 122 \\ 
     \%Blank & 70.6\%  & 54.3\% & 84.3\% & 0.5\% & 3.6\%\\
    \bottomrule
  \end{tabular}
  \caption{The percentage of blank (all zero) saliency maps for each of the 5 neurons with the highest absolute KS-values (compared to group 50) in the Dense-128 layer. The bold neuron numbers are the neurons qualitatively identified as encoding digit 5. We observe that the neurons encoding digit 5 have predominantly larger percentages of blank saliency maps.}
  \label{tab:blankSaliency}
\end{table}


\section{Discussion and Conclusion}
\label{sec:disc-concl}

For the quantitative approach to handling failure modes we could see significant improvement even using quite simplistic classifiers for constructing a correction layer: an increase by almost 20\%pt, avoiding corrections on almost all uncorrupted images was seen in both the linear separation methods: both with logistic regression and SVM. 

On the qualitative side, an inspection of the saliency maps -- see Figure~\ref{fig:false5images} for a selection of particularly illustrative maps, and the supplementary material for a full collection -- showed us that the groups were distinguished from group 50 containing correctly predicted digits 5 differed in network activity either by an activation tuned to detecting 5s, or in an activation that often looked for closing loops and found them in the added noise.  Blank saliency maps were common for the 5-detecting neurons, as can be seen in Table~\ref{tab:blankSaliency}, overwhelmingly so for the groups 40 and 47 where correct predictions were rare, and much less commonly in group 30 where as can be seen in Figure~\ref{fig:Group40distribution} a correct prediction still came with significant strength in the softmax layer.

Using \FiFa on a CNN-based MNIST digit classifier that had to cope with severely corrupted MNIST images we were able to find 39 distinct failure modes based on activations in the antepenultimate and penultimate layers of our CNN model. When inspecting the digit 5 in particular, we found that the three identified failure modes could be distinguished from the wellbehaved parts of input space by specific activations that seemed to code for features corresponding closely to the kinds of misclassifications that were observed.

In addition to inspecting examples, we explored the addition of a correction layer to the CNN model. 
The failure modes act as seeds for training a classifier. The classifier can assign new data to a known failure mode, so that the correction layer can adjust for known behaviour of that failure mode.
For regression models, our suggestion would be to treat the prediction error as bias, and subtract the mean prediction error for the identified failure mode from the model prediction. In the CNN on corrupted MNIST example we use to illustrate the methodology, we impose the ground truth digit from which the identified failure mode emerged as a replacement prediction. By doing this, we could observe up to a 19.33\%pt improvement in prediction accuracy on corrupted data while accidentially including only 0.32\% of uncorrupted observations in the correction groups. The percentage of clean data is in well accordance with that in the failure mode groups; 0.21\%.

\FiFa is generically applicable. While developing the method we have used it to analyze an energy based regression model used to predict temperatures in electric arc steel furnaces. In that application, we found failure modes that consistently over-predicted and under-predicted by close to $100ºC$ . Adjusting the regression by the mean prediction error of the failure group provided significant improvement in the energy model and a qualitative analysis of the failure modes uncovered metallurgically important observations about material composition related to high prediction error.

The \FiFa method picks out high prediction error regions from input space of an arbitrary predictive process, and classifies failure modes that are internally similar but that have significant separation either in the predictive behaviour of the process or in the distance measure of input space. 
Having identified failure modes we can view them as witnesses for misbehaviour in different ways, and produce correspondingly different developments of the predictive process. On the one hand, a failure mode witnesses a region of input space with local bias to the predictive process, and we can correct specifically for that bias by classifying new data as belonging to that failure mode (or not) and correct predictions for the failure mode members. On the other hand the failure mode is a witness for some coherent collection of predictive failures. By inspecting features of input space that distinguish these from other parts of input space we can gain insights about types of failure that could be handled by adjusting the design of the predictive process itself.

%
%
%

\bibliography{mapper-regression}
\bibliographystyle{icml2018}

\end{document}